\documentclass[11pt]{article}
\usepackage{graphicx} % Required for inserting images
\usepackage{amsmath}
\usepackage{amssymb}
\usepackage{booktabs}
\usepackage{geometry}
\usepackage{multirow} 
\usepackage{subcaption}
\usepackage{float}
\usepackage[colorlinks=true, linkcolor=blue, citecolor=blue, urlcolor=blue]{hyperref}
\usepackage{tikz}
\usetikzlibrary{quantikz2}
\title{A Hybrid Quantum Circuit Born Machine Framework for Financial Volatility Forecasting: Quantum-Assisted Training and Classical Inference}
\author{Yixiong Chen}
\date{}

\begin{document}

\maketitle

\begin{abstract}
Accurate financial volatility forecasting is crucial but challenged by the non-linear, highly correlated nature of market data. Recently, quantum computing has emerged as a promising paradigm for solving complex high-dimensional sampling problems. To harness this, we propose a novel hybrid framework combining the temporal representation power of classical neural networks with the distribution-learning capabilities of quantum models. Specifically, we integrate a Long Short-Term Memory (LSTM) network with a Quantum Circuit Born Machine (QCBM). The LSTM extracts dynamic features, while the QCBM acts as a learnable generative prior modeling complex market distributions to guide forecasting. Evaluated on 5-minute high-frequency data from the SSE Composite and CSI 300 indices, our model significantly outperforms a classical LSTM baseline across MSE, RMSE, and QLIKE metrics. Furthermore, by introducing a stochastic ``Drop-Prior" mechanism during training, the LSTM implicitly distills structured information from the quantum prior. This establishes a pragmatic paradigm of ``quantum-assisted training with classical-efficient inference", whereby the model retains its quantum-enhanced accuracy even when the quantum module is entirely disabled during deployment. This demonstrates a practical pathway for leveraging quantum computing to enhance classical models without real-time quantum inference latency.
\end{abstract}

\section{Introduction}
\subsection{Background of Volatility Forecasting}
Financial market volatility, which measures the magnitude of price fluctuations of an asset over a given period, is a fundamental concept in finance. It serves not only as a key input in risk management—directly influencing derivative pricing models such as the Black–Scholes framework—but also plays a critical role in portfolio construction and dynamic asset allocation. However, financial time series often exhibit complex features, including leptokurtosis, volatility clustering, and rapid responses to new information, which make accurate volatility forecasting particularly challenging. Consequently, precise volatility prediction has long been a central objective in both academic research and industry practice.

\subsection{Classical and Deep Learning Methods for Volatility Forecasting}
Classical time series models, such as the  Generalized Autoregressive Conditional Heteroskedasticity (GARCH) model proposed by Bollerslev and its variants \cite{bollerslev1986generalized}, have long been the cornerstone of volatility forecasting. These models can effectively capture the phenomenon of volatility clustering in financial time series. However, they usually rely on strict statistical assumptions and have limited ability to capture the complex non-linear dynamics caused by major market events.

In recent years, with the improvement of computing power and the emergence of massive data, deep learning methods, due to their powerful non-linear modeling capabilities, have shown great potential in the field of volatility forecasting, gradually compensating for the shortcomings of traditional statistical models \cite{zhao2024garch}. Recent review works, such as \cite{ge2022neural}, \cite{gunnarsson2024prediction}, and \cite{sezer2020financial}, have systematically reviewed the application of deep learning in financial forecasting, emphasizing its advantages over traditional statistical models.

Among the many deep learning models, Long Short-Term Memory (LSTM), as a special type of Recurrent Neural Network (RNN), can effectively learn the long-term dependencies in time series data, making it a powerful tool for volatility forecasting \cite{hochreiter1997long, fischer2018deep}. Meanwhile, Convolutional Neural Networks (CNNs), originally designed for image recognition, have also been shown to be effective in extracting local patterns and features from financial time series \cite{hoseinzade2019cnnpred}. Deep Belief Networks (DBNs), as a generative model, can capture the high-dimensional complex distribution of data through unsupervised learning \cite{kuremoto2014time}. Recently, the Transformer model, which has achieved great success in the field of natural language processing, has also been introduced into financial forecasting due to its powerful self-attention mechanism, to capture the complex dependencies between arbitrary positions in a sequence \cite{vaswani2017attention, ramos2021multi}. To combine the advantages of different models, researchers have also proposed a variety of hybrid models, such as combining GARCH with LSTM, using the former to capture the statistical characteristics of volatility and the latter to learn the non-linear patterns in the residuals, thereby significantly improving forecasting accuracy \cite{kim2018forecasting}. In addition, besides traditional data such as price and volume, the use of ``alternative data" such as news text and social media sentiment for forecasting has also become a popular research direction. Text sentiment indicators extracted through Natural Language Processing (NLP) techniques can provide a new dimension of information for volatility forecasting \cite{bollen2011twitter, hu2018listening}. Although deep learning methods have achieved significant success, they also face challenges such as poor model interpretability, sensitivity to hyperparameters, and a tendency to overfit. Therefore, how to design deep learning models that are both accurate and robust, and how to combine them with financial theory, remains an important direction for future research.

\subsection{Quantum Machine Learning and Its Application in Volatility Forecasting}
Meanwhile, quantum computing leverages fundamental principles of quantum mechanics, including superposition and entanglement, to process information, thereby offering the potential to surpass classical computing for certain specific problems. Quantum Machine Learning (QML) is an interdisciplinary field of quantum computing and machine learning that aims to develop learning algorithms that can run on quantum computers to solve complex problems that are difficult for classical methods to handle \cite{biamonte2017quantum}. On this basis, QML has been extended to multiple application areas. For example, in the field of computer vision, Quantum Convolutional Neural Networks (QCNNs) have been proposed for image classification tasks, demonstrating advantages in feature extraction \cite{cong2019quantum}; in Natural Language Processing (NLP), researchers use quantum circuits to simulate the grammatical and semantic structures of language to solve problems such as text classification \cite{coecke2020foundations}; and in the field of time series forecasting, quantum algorithms have also been explored to capture complex patterns and long-term dependencies in sequential data \cite{kaur2021quantum}.

In recent years, researchers have begun to actively explore the application of QML to financial volatility forecasting. The volatility of financial markets often exhibits complex characteristics such as non-linearity, non-Gaussian distribution, and volatility clustering which pose challenges for classical models (such as ARIMA). QML models, due to their unique computational paradigm, are considered to have the potential to capture these complex market dynamics, and the latest benchmark studies have also confirmed their advantages over classical methods \cite{ahmad2026quantum, patel2025advancing, srivastava2023potential}. Current research covers a variety of advanced quantum architectures. In terms of convolutional networks, Quantum Temporal Convolutional Neural Networks (QTCNNs) have been proposed for handling cross-sectional stock return prediction, demonstrating their ability to extract temporal features \cite{chen2025quantum}. As a mainstream architecture for time-series forecasting, LSTM networks exhibit a natural compatibility with quantum computing frameworks. Consequently, various hybrid approaches—such as BLS-QLSTM and other quantum-enhanced LSTM models—have been proposed to improve predictive performance in stock index and asset price forecasting tasks. \cite{su2025bls, kea2024hybrid, arora2025hybrid}. In addition, to capture more complex market dependencies, researchers have also introduced the Quantum Deep Q-Network with Attention (QADQN) \cite{dutta2024qadqn} and contextual quantum neural networks \cite{mourya2026contextual}. In addition to single models, hybrid quantum-classical ensemble learning \cite{weinberg2025hybrid} and regression-type hybrid neural networks \cite{choudhary2025hqnn} have also been shown to be effective in revealing the intrinsic relationships of the market. These studies have jointly promoted the application of QML in financial analysis and paved the way for building more powerful forecasting models as quantum hardware continues to develop.

It is worth emphasizing that in the current ``Noisy Intermediate-Scale Quantum" (NISQ) era, almost all of the above studies have adopted a Hybrid Quantum-Classical computing paradigm. This approach has emerged as the mainstream methodology to mitigate the hardware limitations of NISQ devices—namely, restricted qubit counts and short coherence times—while maximizing quantum computational potential. Specifically, the hybrid framework establishes a complementary division of labor: classical computers manage large-scale data preprocessing, network optimization, and control flows, whereas quantum processors are exclusively tasked with core operations, including high-dimensional feature extraction and complex probability distribution modeling \cite{bharti2022noisy, cerezo2021variational}. Therefore,  the hybrid paradigm is widely regarded as an essential stepping stone toward fault-tolerant quantum computing. Foundational works such as the Variational Quantum Classifier (VQC) \cite{havlivcek2019supervised}, Quantum Generative Adversarial Networks (QGAN) \cite{lloyd2018quantum}, and hybrid quantum neural networks \cite{mitarai2018quantum} exemplify this design and highlight its broad applicability.

Although hybrid quantum-classical algorithms show great potential, they still face several severe challenges in their current stage of development \cite{bharti2022noisy}. First, the high cost of sampling is a key factor limiting their application. The probabilistic nature of quantum measurement requires a large number of repeated measurements (i.e., Shots) for each gradient estimation, resulting in a very high computational cost for parameter updates \cite{schuld2019evaluating}, a problem that is particularly severe when training with real quantum hardware. Second, the data loading process presents a significant hurdle. Constructing quantum embeddings to map classical data into the Hilbert space often involves substantial resource consumption and circuit complexity \cite{lloyd2020quantum}. Furthermore, the choice of encoding strategy directly constrains the model's expressive power and generalization bounds \cite{schuld2021effect, caro2021encoding}, potentially offsetting the acceleration advantages offered by quantum computing. Finally, the gradient coupling between quantum and classical modules creates a severe bottleneck during the optimization process. Since the gradients of both components are intrinsically linked via the chain rule of backpropagation \cite{mitarai2018quantum}, the training stability is heavily compromised by the quantum landscape, which is often plagued by hardware noise and the ``barren plateau" phenomenon \cite{mcclean2018barren, wang2021noise}. This strong dependency leads to a twofold dilemma: the noise and vanishing gradients from the quantum circuit propagate to the entire network, potentially destabilizing the classical convergence; simultaneously, the classical parameter updates are forced to wait for the computationally expensive quantum gradient estimation. Consequently, the quantum module effectively acts as a ``bottleneck layer," severely restricting the overall training efficiency and scalability of the hybrid model.

\subsection{Our Work}
To address the above challenges, we propose a novel hybrid quantum-classical algorithm framework based on the Quantum Circuit Born Machine (QCBM).
The Quantum Circuit Born Machine (QCBM) is a class of generative models based on parameterized quantum circuits. It generates samples by measuring quantum states, and the output probability distribution is determined by the circuit parameters, naturally following the Born rule of quantum mechanics. Thanks to the exponential representation power of quantum states in Hilbert space, QCBM can characterize complex multivariate joint probability distributions with a relatively compact parameter structure. In theory, it has the potential to express distributions that are difficult for classical computational models to simulate efficiently, thus showing unique advantages in high-dimensional correlation scenarios such as financial modeling.

In the field of quantum machine learning, QCBM has been established as an important fundamental generative framework. Studies have shown its significant effectiveness in learning multivariate joint probability distributions \cite{benedetti2019generative}. To address the problems of vanishing gradients and optimization difficulties during training, a differentiable learning method based on the parameter shift rule was proposed, which significantly improved training efficiency and convergence stability \cite{liu2018differentiable}. Further theoretical analysis indicates that, compared to the classical Restricted Boltzmann Machine (RBM), QCBM has higher parameter efficiency in terms of expressive power, being able to approximate more complex target distributions with fewer degrees of freedom \cite{du2020expressive}. At the financial application level, previous work has explored using QCBM to simulate the joint distribution of asset prices and option pricing problems, verifying its potential advantages in characterizing complex correlation structures and tail risks \cite{coyle2021quantum}.

Based on the above research, this paper proposes a hybrid volatility prediction model that combines QCBM with a Long Short-Term Memory (LSTM) network. The core idea of this model is to introduce QCBM as a learnable probability prior generator to model the implicit joint distribution structure of the market, and embed the distribution information it generates into the time series prediction framework of LSTM, thereby enhancing the model's ability to characterize the non-linear and higher-order correlation features of the financial market. Specifically, our model architecture consists of two main parts. The first is the classical backbone network, a standard LSTM network responsible for processing input time series features (e.g., log returns and trading volume)  and extracting their dynamic evolution patterns. The second is the quantum prior module, a QCBM trained to generate a task-related prior probability distribution. Bitstrings sampled from this QCBM are encoded into a prior embedding vector. In the prediction phase, the final hidden state of the LSTM is fused with the prior embedding vector generated by the QCBM through a weighted sum, and then the final volatility prediction is output through a fully connected layer. By alternately training with the LSTM model, the QCBM can learn an optimal prior distribution to guide the LSTM's prediction, helping it to better capture the complex structure in the data, especially during periods of violent market fluctuations. This hybrid method combines the powerful feature extraction capabilities of deep learning with the excellent distribution representation capabilities of quantum models.

This hybrid framework has several significant advantages, effectively avoiding some of the bottlenecks of traditional hybrid models. First, the model adopts a separate training strategy, decoupling the quantum and classical modules. Under this mechanism, the two modules can be optimized independently without complex gradient interaction calculations, thus simplifying the training process and improving efficiency. Second, the optimization of QCBM can use gradient-free optimization algorithms, such as the Constrained Optimization By Linear Approximation (COBYLA) algorithm. These algorithms are more robust to noise, making them more feasible on current NISQ hardware. Third, the decoupled training mechanism fundamentally solves the data loading bottleneck problem. Since there is no need to encode classical time series data into quantum states, the model avoids high data encoding overhead. Finally, the goal of QCBM is to learn the global probability distribution of the data, rather than fitting individual data points.  Consequently, its optimization cost does not scale with the number of training examples, as the training procedure focuses on distributional matching instead of explicit data traversal. This avoids the substantial computational overhead typically associated with sample-wise processing in classical neural networks and conventional hybrid quantum–classical models, thereby demonstrating improved scalability.

Beyond architectural efficiency, we introduce a stochastic ``Drop-Prior" mechanism during the training phase. By randomly deactivating the quantum prior, the classical LSTM backbone is forced to navigate an information bottleneck \cite{tishby2015deep}, effectively internalizing the structured probability distributions generated by the QCBM into its own weight matrices. This process, mirroring implicit knowledge distillation \cite{hinton2015distilling}, ensures that the classical component inherits the quantum-derived structural biases, allowing it to maintain high performance even when the quantum module is absent during deployment.

The main contributions of this paper are as follows:
\begin{itemize}
    \item We propose a novel hybrid quantum-classical framework that innovatively uses a Quantum Circuit Born Machine (QCBM) as a learnable generative prior to enhance the performance of classical prediction models. We apply this framework to the volatility prediction of financial high-frequency data and build the LSTM-QCBM model. To the best of our knowledge, this is the first hybrid architecture to incorporate a learnable quantum prior provided by the QCBM into a supervised learning task, opening up a new avenue for addressing the complex dynamics of financial time series
    \item To effectively train this hybrid model, we propose an alternating training strategy that successfully decouples the optimization processes of the classical and quantum modules. Under this strategy, we first fix the prior generated by the QCBM to enhance the input of the LSTM, allowing us to train the classical model. Subsequently, we fix the trained classical LSTM model and use a well-designed scoring function to guide the QCBM in learning an optimal prior distribution highly relevant to the task. This separate optimization not only avoids the complexity of gradient coupling and backpropagation found in traditional hybrid models but also allows us to use gradient-free optimization algorithms for the quantum module, significantly improving training efficiency and robustness in noisy environments.
    \item We validate the effectiveness and superiority of the proposed architecture through a rigorous empirical analysis. Comprehensive experiments were conducted on two representative high-frequency datasets from the Chinese financial market: the Shanghai Stock Exchange (SSE) Composite Index and CSI 300 Index. The results demonstrate that the proposed LSTM-QCBM significantly outperforms the standard classical LSTM baseline, achieving statistically robust improvements across multiple key evaluation metrics, including Mean Squared Error (MSE), Root Mean Squared Error (RMSE), and Quasi-likelihood (QLIKE) loss. These findings underscore the practical utility of our framework, confirming that the incorporation of a learnable quantum prior effectively guides the classical model in capturing the complex non-linear dynamics and volatility clustering characteristics inherent in financial markets.
    \item We report the experimental discovery of ``Classical-Only Inference", where the hybrid model maintains or even improves its predictive performance when the quantum prior is completely removed during the testing phase. We link this phenomenon to the Learning Using Privileged Information (LUPI) framework \cite{vapnik2009new,vapnik2015learning}, demonstrating that the QCBM acts as a pedagogical guide during the training process. This discovery provides a pragmatic strategy for deploying quantum-enhanced models: utilizing expensive quantum resources for training to reshape the classical optimization trajectory, while relying on classical-level efficiency for real-world inference.
\end{itemize}

\section{Methodology}
\subsection{Quantum Circuit Born Machine (QCBM)}
The Quantum Circuit Born Machine (QCBM) is a quantum generative model built upon a parameterized quantum circuit (PQC).  In contrast to classical generative models that rely on thermal distributions, such as Boltzmann machines, QCBM directly uses the wave function of
a quantum state to represent a probability distribution and directly generates samples by per-
forming projective measurements on qubits, which is theoretically more efficient than classical
methods such as Gibbs sampling \cite{ackley1985learning}.

A QCBM prepares a complex final state $|\psi(\boldsymbol{\theta})\rangle$ by applying a series of learnable quantum gates, controlled by parameters $\boldsymbol{\theta}$, to an initial quantum state, typically $|0\rangle^{\otimes n}$. According to the Born rule of quantum mechanics, when measuring this final state in the computational basis, the probability of obtaining a certain bit string $\mathbf{x} \in \{0, 1\}^n$ is:
\begin{equation}
    P_{\boldsymbol{\theta}}(\mathbf{x}) = |\langle \mathbf{x} | \psi(\boldsymbol{\theta}) \rangle|^2
\end{equation}
This probability distribution $P_{\boldsymbol{\theta}}(\mathbf{x})$ is called the Born distribution. The goal of QCBM is to optimize the parameters $\boldsymbol{\theta}$ so that the generated Born distribution approximates a target data distribution as closely as possible. Due to the exponential dimensionality of the quantum state space, where the Hilbert space for $n$ qubits scales as $2^n$, QCBM has the potential to represent highly complex and correlated probability distributions that are difficult for classical models to describe effectively.

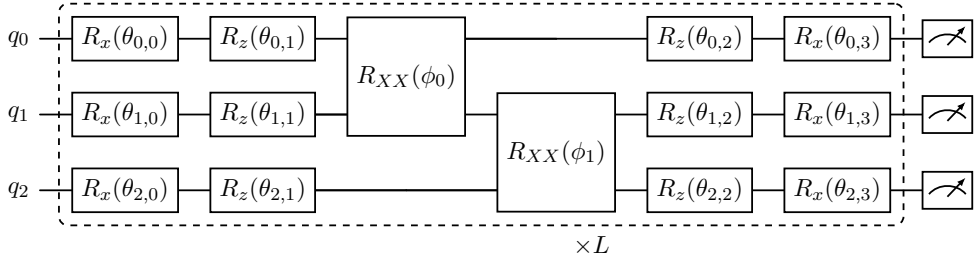
\begin{figure}[htbp]
    \centering
    \scalebox{0.85}{
    \begin{quantikz}
    \lstick{$q_0$} & \gate{R_x(\theta_{0,0})} \gategroup[wires=3, steps=6, style={dashed, rounded corners, inner sep=2pt}, label style={label position=below right, xshift=-1pt, yshift=-15pt}]{$\times L$}  & \gate{R_z(\theta_{0,1})} & \gate[2]{R_{XX}(\phi_0)} & \qw & \gate{R_z(\theta_{0,2})} & \gate{R_x(\theta_{0,3})} & \meter{} \\
    % --- Qubit 1 ---
    \lstick{$q_1$} & \gate{R_x(\theta_{1,0})} & \gate{R_z(\theta_{1,1})} & \qw & \gate[2]{R_{XX}(\phi_1)} & \gate{R_z(\theta_{1,2})} & \gate{R_x(\theta_{1,3})} & \meter{} \\
    % --- Qubit 2 ---
    \lstick{$q_2$} & \gate{R_x(\theta_{2,0})} & \gate{R_z(\theta_{2,1})} & \qw & \qw & \gate{R_z(\theta_{2,2})} & \gate{R_x(\theta_{2,3})} & \meter{} 
    \end{quantikz}
    }
    \caption{Quantum Circuit Born Machine Structure}
    \label{fig:qcbm_structure}
\end{figure}

In our work, we employ a QCBM with a layered architecture, as depicted in Figure \ref{fig:qcbm_structure}.
Specifically, each layer first applies $R_x$ and $R_z$ rotation gates to each qubit, then entangles adjacent qubit pairs through $R_{XX}$ gates, and finally applies another set of $R_z$ and $R_x$ rotation gates. The entire circuit repeats this layered structure L times, and finally measures all qubits to obtain a classical bit string. The parameters of the circuit, including the rotation angles $\theta$ and the $R_{XX}$ gate parameters $\phi$, are trainable. By optimizing these parameters, the QCBM can learn to generate a specific prior distribution.
\subsection{Classical Long Short-Term Memory (LSTM) Network}
The Long Short-Term Memory (LSTM) network, a specialized variant of Recurrent Neural Networks (RNNs), serves as the classical backbone of our hybrid architecture due to its robust capacity for modeling sequential data and mitigating the vanishing gradient problem \cite{hochreiter1997long}. In the context of financial volatility forecasting, the LSTM is tasked with extracting temporal dependencies from historical market data. 

Given an input sequence $X = (\mathbf{x}_1, \mathbf{x}_2, \ldots, \mathbf{x}_T)$, where $\mathbf{x}_t$ represents the financial features at time step $t$, the LSTM updates its hidden state $\mathbf{h}_t$ and cell state $\mathbf{c}_t$ through a series of gating mechanisms. Specifically, the forget gate $\mathbf{f}_t$, input gate $\mathbf{i}_t$, and output gate $\mathbf{o}_t$ control the flow of information:
\begin{align}
\mathbf{f}_t &= \sigma(\mathbf{W}_f \cdot [\mathbf{h}_{t-1}, \mathbf{x}_t] + \mathbf{b}_f) \\
\mathbf{i}_t &= \sigma(\mathbf{W}_i \cdot [\mathbf{h}_{t-1}, \mathbf{x}_t] + \mathbf{b}_i) \\
\tilde{\mathbf{c}}_t &= \tanh(\mathbf{W}_c \cdot [\mathbf{h}_{t-1}, \mathbf{x}_t] + \mathbf{b}_c) \\
\mathbf{c}_t &= \mathbf{f}_t \odot \mathbf{c}_{t-1} + \mathbf{i}_t \odot \tilde{\mathbf{c}}_t \\
\mathbf{o}_t &= \sigma(\mathbf{W}_o \cdot [\mathbf{h}_{t-1}, \mathbf{x}_t] + \mathbf{b}_o) \\
\mathbf{h}_t &= \mathbf{o}_t \odot \tanh(\mathbf{c}_t)
\end{align}
where $\sigma$ denotes the sigmoid activation function, $\odot$ represents element-wise multiplication, and $\mathbf{W}$ and $\mathbf{b}$ are the learnable weight matrices and biases, respectively. The final hidden state $\mathbf{h}_T$, which encapsulates the extracted temporal features of the entire sequence, is then passed to a fully connected  layer to generate the final volatility prediction $\hat{y}$.

\subsection{QCBM-based Hybrid Quantum-Classical Algorithm}
\begin{figure}[ht]
    \centering
    \includegraphics[width=0.8\textwidth]{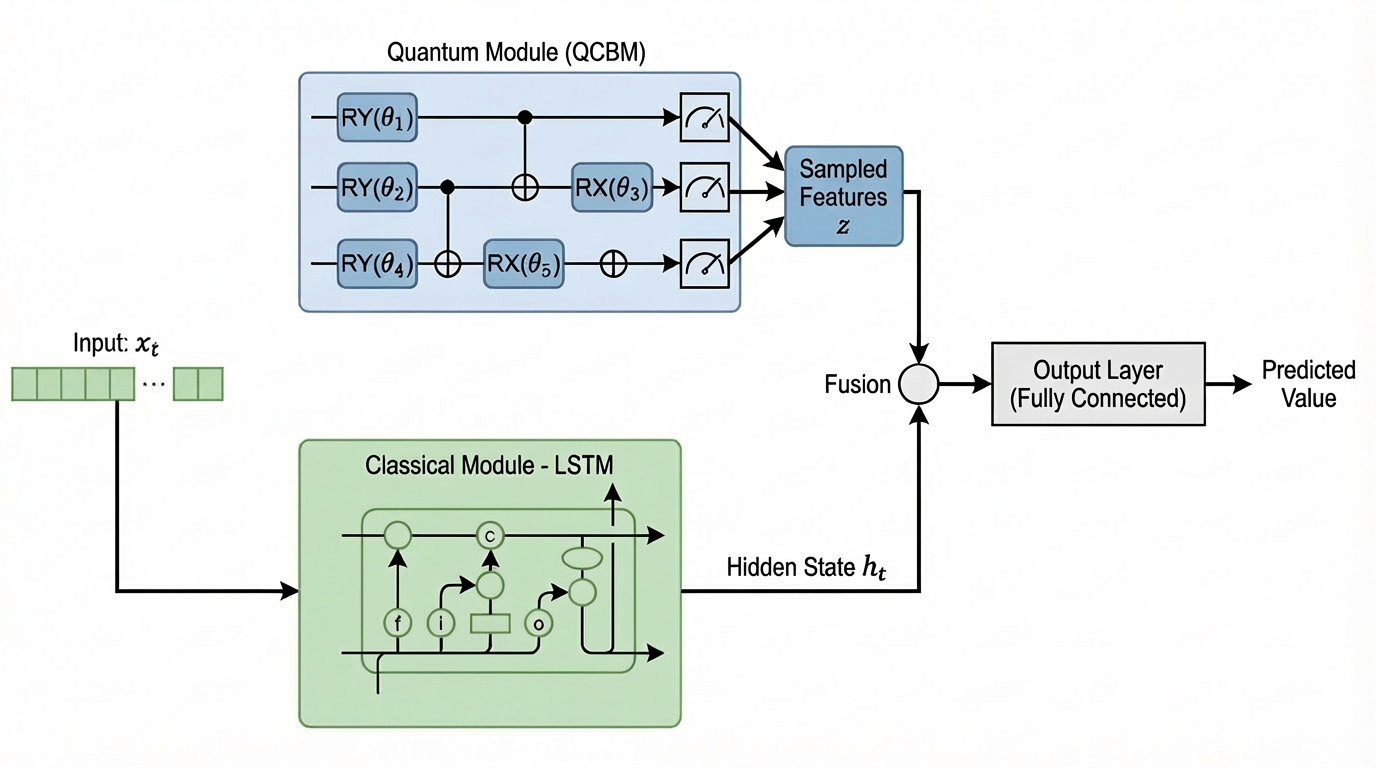} % A real image needs to be inserted manually
    \caption{Schematic diagram of the LSTM-QCBM hybrid model architecture.}
    \label{fig:architecture}
\end{figure}

The hybrid prediction model we propose, namely LSTM-QCBM, organically combines a classical deep learning model with a quantum generative model. Its overall architecture is shown in Figure \ref{fig:architecture}.
The model consists of two core modules:
\begin{itemize}
    \item LSTM backbone network. This module serves as the core classical feature extractor for modeling sequential dependencies in time-series data. The input is a multi-dimensional feature sequence $X = (x_1, x_2, \dots, x_T)$ containing information such as historical prices and trading volumes. The LSTM network captures the long-term dependencies in the sequence through its internal gating mechanism and outputs the hidden state of the last time step $\mathbf{h}_T$
    \begin{equation}
        \mathbf{h}_T = \text{LSTM}(X; W_{\text{lstm}})
    \end{equation}
    where $W_{\text{lstm}}$ represents the learnable parameters of the LSTM network.

    \item QCBM prior module. The core of this module is an $n$-qubit QCBM. During the training process, the QCBM is optimized to generate a prior distribution that is beneficial for the prediction task. By sampling from the trained QCBM, a bit string $\mathbf{z} \in \{0, 1\}^n$ can be obtained. This bit string is mapped to an embedding vector $\mathbf{e}_{\text{prior}}$ with the same dimension as the LSTM hidden state through a projection layer (i.e., a fully connected layer)
    \begin{equation}
        \mathbf{e}_{\text{prior}} = \text{Linear}(\mathbf{z}; W_{\text{proj}}),
    \end{equation}
    where $W_{\text{proj}}$ are weight parameters.
\end{itemize}
The final predicted value is obtained by fusing the hidden state of the LSTM and the prior embedding of the QCBM. The fusion method is a weighted sum:
\begin{equation}
    \mathbf{h}_{\text{fused}} = \mathbf{h}_T + \alpha \cdot \mathbf{e}_{\text{prior}}
\end{equation}
where $\alpha$ is a hyperparameter used to control the strength of the quantum prior. The fused vector $\mathbf{h}_{\text{fused}}$ is then passed through an output layer (i.e., a fully connected layer) to obtain the final volatility prediction value $\hat{y}$.
\begin{equation}
    \hat{y} = \text{Linear}(\mathbf{h}_{\text{fused}}; W_{\text{out}}),
\end{equation}
where $W_{\text{out}}$ denotes the learnable weight matrix of the output layer.

To make the two modules work together, we designed an alternating training strategy. The entire training process is divided into two alternating phases:
\begin{enumerate}
    \item Training the LSTM-QCBM model. In this phase, the parameters $\boldsymbol{\theta}$ of the QCBM are fixed. We use the prior samples generated from the current QCBM, together with the input classical time series data, to train the parameters of the entire hybrid model, including the LSTM, projection layer, and output layer. The training objective is to minimize the loss function between the predicted value and the true value, such as the mean squared error (MSE)
    \begin{equation}
        \mathcal{L}_{\text{model}} = \frac{1}{N} \sum_{i=1}^N (\hat{y}_i - y_i)^2
    \end{equation}

    \item Optimizing the QCBM. In this phase, the parameters of the hybrid model are fixed. We first calculate a score for each unique prior sample generated by the current model. This score reflects the contribution or fitness of the prior sample to the model's prediction. The larger the score, the greater the contribution or fitness. In our implementation, the score is defined as some transformation of the model's prediction performance metric when using that prior sample. For example, we can use the negative of the model's mean squared error (MSE) to calculate the score. At this time, for a prior sample $\mathbf{z}_j$, its score is $s_j = -\text{MSE}(\mathbf{z}_j)$.
    Then, we use the Softmax function to transform these scores into a target probability distribution $P_{\text{target}}$:
    \begin{equation}
        P_{\text{target}}(\mathbf{z}_j) = \frac{e^{ s_j}}{\sum_k e^{ s_k}}.
    \end{equation}
    This target distribution represents the ideal prior distribution. Finally, we optimize the parameters $\boldsymbol{\theta}$ of the QCBM to minimize the KL divergence between its generated Born distribution $P_{\boldsymbol{\theta}}$ and the target distribution $P_{\text{target}}$.
    \begin{equation}
        \mathcal{L}_{\text{qcbm}} = \text{KL}(P_{\text{target}} || P_{\boldsymbol{\theta}})
    \end{equation}
    Here we use the classical optimizer COBYLA to optimize the parameters.
\end{enumerate}
By alternately executing these two steps in each epoch, the QCBM gradually learns to generate prior information that can guide the LSTM to make more accurate predictions, while the LSTM learns how to most effectively use these quantum priors. This co-evolutionary mechanism is the core advantage of this hybrid algorithm.

The hybrid prediction algorithm based on QCBM that we propose has several significant advantages:
\begin{itemize}
    \item Decoupled training mechanism: The quantum and classical modules employ a separate training strategy, which eliminates the need for gradient backpropagation within the quantum module. This contrasts with many existing hybrid models where parameter gradient calculations are highly coupled due to the chain rule, leading to frequent interactions between modules and thus increasing the training overhead, especially on real quantum hardware. Our method significantly improves training efficiency through decoupled training.
    \item Efficient and robust quantum optimization: Since the optimization of the quantum module does not rely on gradient information from the classical module, we can use gradient-free optimization algorithms (such as COBYLA). Such algorithms not only accelerate the convergence of quantum parameters but also exhibit stronger resistance to noise compared to gradient-based optimizers, thereby showing higher robustness on current NISQ devices.
    \item Elimination of the data loading bottleneck: A key challenge in many hybrid quantum algorithms is loading large-scale classical data into quantum states. This process, known as data encoding, often requires a large number of quantum gate operations, and its complexity can grow exponentially with the data dimension, creating a severe performance bottleneck that can even offset the potential advantages of quantum computing. Our framework fundamentally solves this problem through its unique decoupled design. The quantum module (QCBM) acts as a generator whose task is to learn a prior distribution rather than directly processing input data. Therefore, the model completely avoids the costly step of encoding the classical dataset into quantum states.
    \item Scalable optimization process: The training cost of traditional deep learning models is typically proportional to the size of the dataset, as they need to repeatedly iterate through data samples to calculate losses and update gradients. In contrast, the QCBM optimization process in our framework offers excellent scalability. Its goal is to learn a global prior distribution that describes the overall characteristics of the data, rather than fitting each individual data point. Consequently, its optimization cost depends mainly on the number of generated samples and the complexity of the model parameters, and is independent of the total number of samples in the training dataset. This ``data-traversal-free" characteristic gives the framework a significant computational efficiency advantage over traditional methods when dealing with massive datasets.
    \item High flexibility: The proposed framework is characterized by a high degree of flexibility. First, its scoring function is highly customizable, allowing us to incorporate diverse domain knowledge or business metrics (e.g., different risk or return measures) and inject this knowledge into the classical model through prior samples to enhance its learning and generalization capabilities. Second, the structure of the quantum module, inclcuding the number of qubits and circuit depth, can be freely scaled according to the complexity of the task and available quantum resources. Finally, the integration of quantum priors is adaptable: they can be introduced at various stages of the classical network (input, hidden, or output layers) using different fusion mechanisms (e.g., summation or concatenation), offering a broad design space for architectural exploration.
\end{itemize}

\section{Experiments}
\subsection{Data and Evaluation Metrics}
\subsubsection{Dataset}
We used two major indices of the Chinese A-share market as experimental data: the Shanghai Stock Exchange (SSE) Composite Index and CSI 300 Index. The data spans from November 1, 2025, to January 1, 2026, with a sampling frequency of 5 minutes. For each time point, the data features used include open, close, high, low, and volume.

Following standard practices in financial time series analysis, we performed a series of feature engineering steps. First, we calculated the log returns as the primary measure of price changes. Then, we calculated the realized volatility as the prediction target. Specifically, our task is to use information from a past period (e.g., the most recent 50 minutes, corresponding to 10 time steps) to predict the realized volatility of the next 25 minutes (i.e., 5 time steps). The realized volatility is calculated as the standard deviation of the log returns over the next 5 time steps.

\subsubsection{Evaluation Metrics}
To comprehensively evaluate the predictive performance of the models, we adopted three commonly used metrics in the field of volatility forecasting:
\begin{enumerate}
    \item Mean Squared Error (MSE):
    \begin{equation}
        \text{MSE} = \frac{1}{N} \sum_{i=1}^N (\hat{y}_i - y_i)^2
    \end{equation}
    \item Root Mean Squared Error (RMSE):
    \begin{equation}
        \text{RMSE} = \sqrt{\frac{1}{N} \sum_{i=1}^N (\hat{y}_i - y_i)^2}
    \end{equation}
    \item QLIKE Loss Function: This is a loss function commonly used in GARCH models and volatility forecasting, which penalizes errors in low-volatility regions more heavily.
    \begin{equation}
        \text{QLIKE} = \frac{1}{N} \sum_{i=1}^N \left( \log(\hat{y}_i) + \frac{y_i}{\hat{y}_i} \right)
    \end{equation}
\end{enumerate}
where $y_i$ is the true realized volatility, and $\hat{y}_i$ is the model's predicted value. For all metrics, a lower value indicates better model performance.

\subsection{Experimental Setup}
\subsubsection{Model Architecture}
\begin{itemize}
    \item LSTM-QCBM: The classical part consists of a two-layer LSTM network with a hidden dimension of 32. The quantum module employs a 12-qubit QCBM architecture with a depth of 3 layers, corresponding to $L=3$ in Figure \ref{fig:qcbm_structure}. The fusion weight $\alpha$ for the quantum prior is set to 0.5.
    \item Baseline Model: For comparison, we implemented a purely classical LSTM model. Its architecture is identical to the classical part of LSTM-QCBM model but does not include any QCBM module.
\end{itemize}

\subsubsection{Training Hyperparameters}
The dataset was partitioned chronologically into training, validation, and test sets with proportions of 70\%, 20\%, and 10\%, respectively. The model was trained using the Adam optimizer with a learning rate of 1e-3 and a batch size of 64. The total number of training epochs was 300. In each epoch, the QCBM was optimized for 50 iterations using the COBYLA optimizer. To ensure the robustness and stability of our results, all experiments are repeated with three different random seeds. We report the mean and standard deviation of each metric over these runs.

\subsection{Results and Analysis}
\subsubsection{Quantitative Results}\label{sec:quantitative_results}
We evaluated the performance of the LSTM-QCBM model and the purely classical LSTM baseline model on the test set. Table \ref{tab:results} shows the comparison of the two models on the three evaluation metrics.
\begin{table}[ht]
    \centering
    \caption{Performance comparison of models on the test set. All metrics are the lower the better.}
    \label{tab:results}
    \small
    \begin{tabular}{@{}lcccc@{}}
        \toprule
        Dataset & Model & MSE ($\times 10^{-6}$) & RMSE ($\times 10^{-3}$) & QLIKE \\
        \midrule
        \multirow{2}{*}{SSE Index} & LSTM  & $5.000 \pm 4.000$ & $2.076 \pm 0.898$ & $-2.073 \pm 0.433$ \\
        & \textbf{LSTM-QCBM } & $\mathbf{3.000 \pm 1.000}$ & $\mathbf{1.659 \pm 0.264}$ & $\mathbf{-4.421 \pm 0.488}$ \\
        \midrule
        \multirow{2}{*}{CSI 300} & LSTM  & $4.000 \pm 2.000$ & $1.959 \pm 0.582$ & $-3.223 \pm 0.382$ \\
        & \textbf{LSTM-QCBM } & $\mathbf{3.670 \pm 1.470}$ & $\mathbf{1.943 \pm 0.515}$ & $\mathbf{-3.842 \pm 0.841}$ \\
        \bottomrule
    \end{tabular}
\end{table}
The LSTM-QCBM model significantly outperforms the purely classical LSTM model on the three core metrics of MSE, RMSE, and QLIKE, showing consistent and stable performance improvement across different market indices. Specifically, on the CSI 300 Index, the hybrid model achieves improvements of 8.25\% in MSE, 0.82\% in RMSE, and 19.21\% in QLIKE compared to the classical baseline. On the SSE Index dataset, the advantage of the quantum-enhanced model is even more pronounced, with substantial improvements of 40.00\% in MSE, 20.09\% in RMSE, and 113.27\% in QLIKE. These results strongly demonstrate the effectiveness of our proposed hybrid quantum-classical framework. We attribute this performance improvement to the powerful distribution learning capability of QCBM, which can capture complex, non-classical statistical correlations in financial time series and provide high-quality prior information to the classical LSTM model. This quantum-enhanced learning mechanism enables the hybrid model to predict market volatility more accurately, especially during critical periods of rapid market dynamics changes.

\subsubsection{Convergence Analysis}
\begin{figure}[ht]
    \centering
     \includegraphics[width=0.7\textwidth]{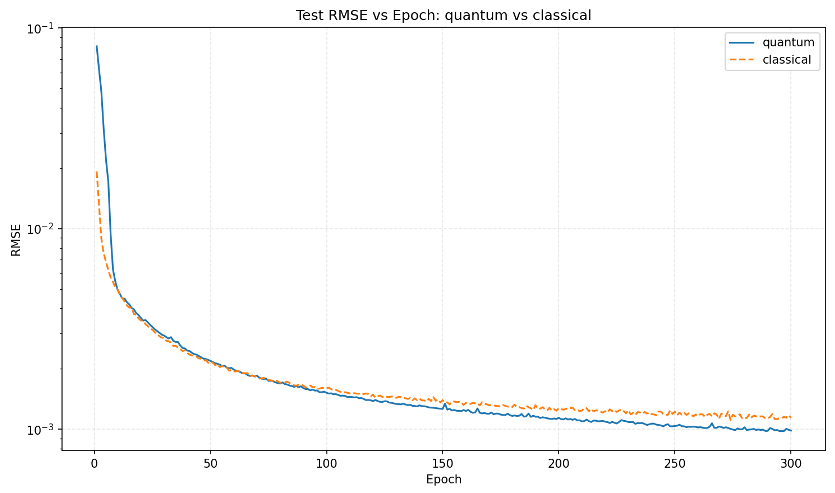} % Actual image needs to be inserted manually
    \caption{Comparison of RMSE convergence curves of the models on the test set.}
    \label{fig:rmse_curve}
\end{figure}

To further investigate the training dynamics of the models, we plotted the RMSE curves of the two models on the CSI 300 test set as a function of the number of training epochs, as shown in Figure \ref{fig:rmse_curve}.
As can be seen from the figure, both models show good convergence trends as training progresses. It is worth noting that in the early stage of training, the classical LSTM model has a lower initial RMSE, but the LSTM-QCBM model, benefiting from the continuous injection of quantum prior knowledge, shows a very fast convergence speed. Its RMSE curve drops rapidly below that of the classical model after a short initial phase. This phenomenon indicates that the introduction of prior knowledge effectively guides the model's learning direction, enabling it to approach the optimal solution more quickly. In the subsequent majority of the training process, the curve of the LSTM-QCBM model is consistently lower than that of the classical model, indicating that it achieves better performance on the test set. In the end, both models converge to a low error level, but the LSTM-QCBM model achieves a significantly lower final RMSE value, proving the effectiveness of the hybrid method in improving prediction accuracy.

\subsubsection{Verification of QCBM's Probability Distribution Learning Ability}
To further verify the effectiveness of the QCBM in our model, we visualized the evolution of the probability distribution it generates on the CSI 300 dataset.

\begin{figure}[htbp!]
    \centering
    \includegraphics[width=0.9\textwidth]{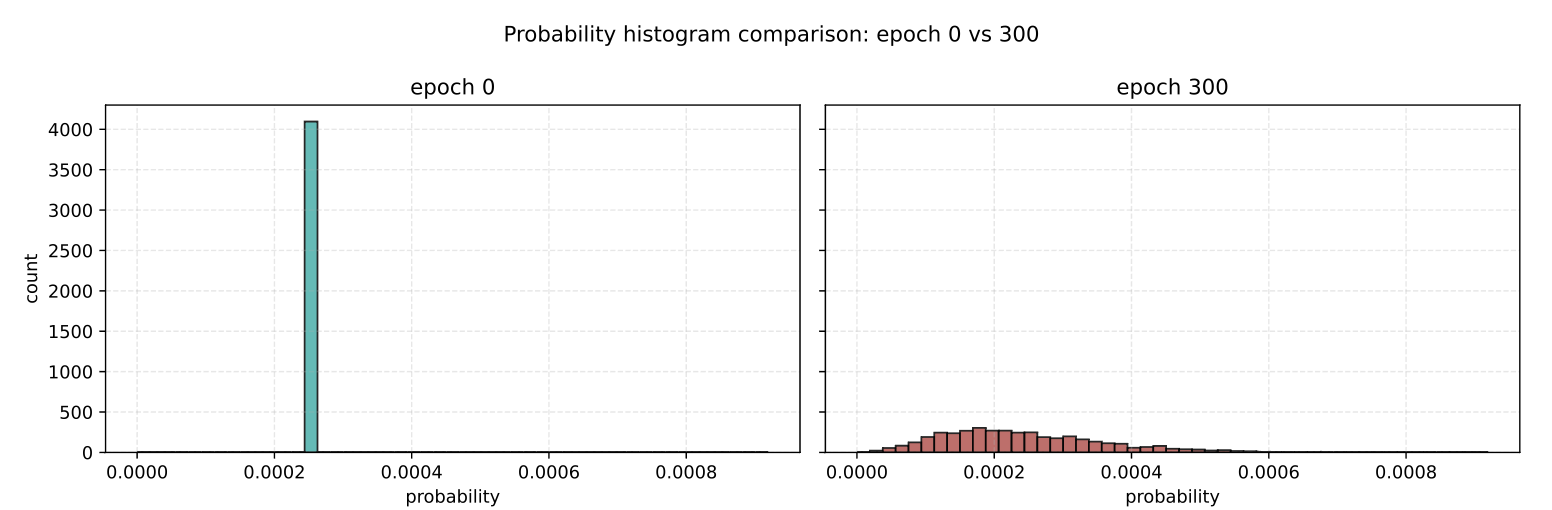}
    \caption{Histogram comparison of the probability distributions generated by the QCBM at epoch 0 and epoch 300.}
    \label{fig:prob_hist}
\end{figure}
First, we examined the change in the probability distribution of the samples generated by QCBM before and after training. As shown in Figure \ref{fig:prob_hist}, prior to training, since the quantum circuit parameters are randomly initialized, the generated probability distribution is approximately uniform, with the probability values of all samples concentrated in a very narrow range. This is consistent with the characteristics of an untrained, uninformative quantum state. However, after 300 epochs of alternating training, the probability distribution shows a significant deviation and a wider spread. This clearly indicates that QCBM has evolved from a random state to a complex model capable of characterizing the inherent non-uniform probability structure of the data.

\begin{figure}[htbp!]
    \centering
    \includegraphics[width=0.5\textwidth]{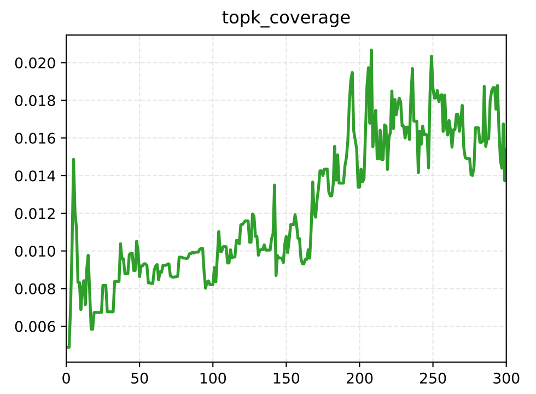}
    \caption{Cumulative probability coverage of the Top-$k$ samples ($k=20$) as a function of training epochs.}
    \label{fig:topk_coverage}
\end{figure}
Second, we confirmed QCBM's learning ability from another perspective. Figure \ref{fig:topk_coverage} tracks the cumulative probability of the Top-$k$ samples with the highest probability as a function of the training epoch. In this analysis, we set $k=20$. We can observe that the Top-20 cumulative probability shows a continuous upward trend as training progresses. This indicates that the quantum circuit is gradually learning to concentrate the probability mass into the "important" sample subspace that contributes more to the model's prediction. In other words, during the training process, the model continuously improves its ability to cover and express the key structures of the target distribution, rather than distributing the probability evenly among all possible outputs.

Based on the above analyses, we conclude that the QCBM module does not function as a static or randomly initialized component; rather, it exhibits dynamic learning behavior during training. Through joint optimization with the classical LSTM module, it learns a task-relevant prior distribution, which contributes to improved predictive performance of the overall hybrid model.

\subsubsection{Classical Only Inference and Implicit Knowledge Distillation}
To rigorously investigate how the QCBM prior interacts with the classical backbone and whether the classical network can internalize this quantum-derived structural information, we designed a novel ``Drop-Prior" mechanism and evaluated the model under a Classical Only Inference setting.

The core concept is to force the classical LSTM network to simultaneously learn to predict with and without the assistance of the quantum prior during training. In each training batch, the model activates the quantum prior by setting $w=0.5$ with a $50\%$ probability and deactivates it by setting $w=0.0$ with a $50\%$ probability. However, when calculating the unique scores for updating the QCBM, the prior is strictly enforced at $w=0.5$ to accurately evaluate the quality of the generated bitstrings. During the final testing phase, we systematically evaluated two distinct inference strategies for the hybrid model: ``With Quantum Prior'' inference utilizing $w=0.5$, which retains the quantum prior, and ``Classical Only'' inference utilizing $w=0.0$, which completely disables it. To establish a rigorous baseline, we directly compared these hybrid strategies against the purely classical LSTM model previously evaluated in Section \ref{sec:quantitative_results}, which was trained entirely without the QCBM module.

\begin{table}[htbp!]
\caption{Comparison of Classical Only Inference and With Quantum Prior Inference against a pure classical baseline on the CSI 300 and SSE Index Datasets. The table highlights the advantage of the stochastic ``Drop-Prior'' mechanism. \textbf{Standard} refers to models trained without Drop-Prior, while \textbf{Drop-Prior} refers to models trained with it. The ``With Quantum Prior'' configuration denotes the hybrid model utilizing the quantum prior during testing ($w=0.5$). In contrast, the ``Classical Only'' configuration represents testing with the quantum prior completely disabled ($w=0.0$).}
\label{tab:zero_weight_inference}
\centering
\resizebox{0.95\textwidth}{!}{%
\begin{tabular}{@{}llccccc@{}}
\toprule
Dataset & Model & Training Strategy & Inference Mode  & MSE ($\times 10^{-6}$) & RMSE ($\times 10^{-3}$) & QLIKE \\
\midrule
\multirow{5}{*}{SSE Index} & LSTM & - & - & $5.000 \pm 4.000$ & $2.076 \pm 0.898$ & $-2.073 \pm 0.433$ \\
\cmidrule{2-7}
& \multirow{4}{*}{QCBM-LSTM} & \multirow{2}{*}{Standard} & With Quantum Prior & $3.000 \pm 1.000$ & $1.659 \pm 0.264$ & $-4.421 \pm 0.488$ \\
& & & Classical Only & $4652.360 \pm 4048.970$ & $56.615 \pm 46.590$ & $-1.421 \pm 0.488$ \\
\cmidrule{3-7}
& & \multirow{2}{*}{Drop-Prior} & With Quantum Prior & $\mathbf{3.000 \pm 2.000}$ & $\mathbf{1.536 \pm 0.661}$ & $-4.536 \pm 0.248$ \\
& & & Classical Only & $\mathbf{3.000 \pm 2.000}$ & $1.560 \pm 0.704$ & $\mathbf{-4.790 \pm 0.235}$ \\
\midrule
\multirow{5}{*}{CSI 300} & LSTM & - & - & $4.000 \pm 2.000$ & $1.959 \pm 0.582$ & $-3.223 \pm 0.382$ \\
\cmidrule{2-7}
& \multirow{4}{*}{QCBM-LSTM} & \multirow{2}{*}{Standard} & With Quantum Prior & $3.670 \pm 1.470$ & $1.943 \pm 0.515$ & $-3.842 \pm 0.841$ \\
& & & Classical Only & $4652.360 \pm 4048.970$ & $56.615 \pm 46.590$ & $-0.842 \pm 0.841$ \\
\cmidrule{3-7}
& & \multirow{2}{*}{Drop-Prior} & With Quantum Prior & $4.000 \pm 3.000$ & $1.975 \pm 0.717$ & $-3.662 \pm 0.793$ \\
& & & Classical Only & $\mathbf{4.000 \pm 3.000}$ & $\mathbf{1.937 \pm 0.746}$ & $\mathbf{-4.761 \pm 0.333}$ \\
\bottomrule
\end{tabular}%
}
\end{table}
To establish a comprehensive comparison, we evaluated the performance on both the CSI 300 and SSE Index datasets across three random seeds. The results are summarized in Table \ref{tab:zero_weight_inference}. As anticipated, under the Standard training strategy, the QCBM-LSTM utilizing the ``With Quantum Prior'' mode, where the fusion weight $w$ is set to $0.5$, significantly outperforms the purely classical LSTM baseline across almost all metrics. This finding is consistent with our observations in Section \ref{sec:quantitative_results}. However, if the quantum prior is forcefully removed from this standard-trained model during the testing phase—corresponding to the ``Classical Only'' mode where $w=0.0$—the predictive performance drastically collapses across all evaluation metrics. Specifically, the MSE experiences an unprecedented deterioration, surging from approximately $3 \times 10^{-6}$ to over $4600 \times 10^{-6}$, representing an error amplification of three orders of magnitude. Concurrently, the RMSE increases drastically to $56.615 \times 10^{-3}$, and the QLIKE metric also degrades significantly. This collapse underscores the severe co-adaptation between the classical and quantum modules during Standard training, rendering the model entirely dependent on the quantum prior for viable inference.

In stark contrast, a highly counter-intuitive yet theoretically significant phenomenon emerges when analysing the Drop-Prior training strategy. The performance of the Drop-Prior trained QCBM-LSTM does not deteriorate upon the complete removal of the quantum prior. Instead, the ``Classical Only'' inference often surpasses or remains highly competitive with the ``With Quantum Prior'' inference, particularly in the QLIKE metric. For instance, on the CSI 300 dataset, the ``Classical Only'' inference achieves relative improvements of $1.92\%$ in RMSE, decreasing from $1.975$ to $1.937$, and $30.01\%$ in QLIKE, improving from $-3.662$ to $-4.761$, compared to its ``With Quantum Prior'' counterpart. We interpret this phenomenon through three theoretical mechanisms:

\begin{itemize}
    \item \textbf{Dynamic Information Bottleneck and Implicit Knowledge Distillation}: In standard joint training utilizing the ``With Quantum Prior'' mode, the LSTM tends to co-adapt with the quantum features. The fifty percent drop-prior mechanism constructs a dynamic information bottleneck, a concept rooted in the Information Bottleneck principle for deep learning \cite{tishby2015deep}. To maintain low loss when the prior is masked, the LSTM is forced to implicitly internalize and distil the structured information inherent in the quantum prior into its own weight matrices, mirroring the dynamics of implicit knowledge distillation \cite{hinton2015distilling}. Consequently, under ``Classical Only'' inference, the resulting classical architecture exhibits significantly enhanced representational capacity compared to standard LSTM models.
    \item \textbf{Quantum Prior as Structured Noise Regularization}: During training, the discrete bitstrings sampled by the QCBM act as highly structured quantum noise, mitigating the overfitting of the LSTM to specific noise patterns and improving overall generalization. However, during deterministic inference, retaining these samples introduces unnecessary variance. By applying ``Classical Only'' inference, the model preserves the generalization benefits derived from low bias while eliminating the sampling-induced random noise associated with high variance. This optimal bias-variance trade-off explains why ``Classical Only'' inference outperforms ``With Quantum Prior'' inference.
    \item \textbf{Learning Using Privileged Information (LUPI)}: This paradigm successfully implements Vapnik's LUPI framework \cite{vapnik2009new,vapnik2015learning}. The quantum module acts as a pedagogical guide, providing privileged information that reshapes the optimization trajectory of the classical network toward flatter and more robust local minima. Discarding the quantum module during deployment achieves classical-level inference efficiency while retaining optimal generalization performance on core financial metrics such as QLIKE.
\end{itemize}

\section{Conclusion and Discussion}
In this paper, we introduce a novel hybrid quantum-classical computing framework designed to enhance the performance of complex tasks by combining classical deep learning models with quantum generative models. The core idea is to utilize the QCBM as a learnable prior generator to provide valuable prior information to the classical model. To validate the effectiveness of this framework, we applied it to the task of financial time series volatility forecasting and designed a specific implementation, the LSTM-QCBM model. This model combines the powerful temporal feature extraction capabilities of LSTM networks with the excellent probability distribution modeling capabilities of QCBM. Through a well-designed alternating training strategy, this model can effectively improve prediction accuracy, demonstrating the potential and practicality of our proposed framework. We conducted extensive experiments on high-frequency datasets of the SSE Index and the CSI 300 Index. The results show that compared to the purely classical LSTM baseline model, our LSTM-QCBM model achieves significant performance improvements on several key metrics such as MSE, RMSE, and QLIKE. This strongly demonstrates the effectiveness of quantum prior information in enhancing the ability of classical machine learning models to handle complex financial data.

This work not only provides a highly effective architecture for volatility forecasting but, more importantly, it establishes a viable and broadly applicable hybrid quantum-classical pathway for the practical deployment of quantum computing across various complex domains. It demonstrates that even in the current NISQ era, real-world challenges can be effectively addressed by strategically integrating quantum resources with classical algorithms.

Crucially, through our extensive evaluation of the ``Classical Only'' inference setting, we have uncovered a novel paradigm for quantum-classical integration. By implementing a stochastic drop-prior mechanism during the training phase, the classical network successfully achieves implicit knowledge distillation and realizes the framework of Learning Using Privileged Information (LUPI). This mechanism ensures that the model can be deployed without the QCBM module during inference while still achieving superior predictive performance. This establishes a highly pragmatic strategy for the deployment of Quantum Machine Learning (QML) technologies: \textit{``Quantum-assisted training with classical-efficient inference.''}

This asymmetric hybrid paradigm elegantly bypasses the most severe bottlenecks of current quantum computing while preserving its core advantages. Primarily, it overcomes the significant deployment barriers associated with physical quantum hardware. Because current quantum computers are prohibitively expensive, environmentally fragile (e.g., requiring cryogenic cooling), and typically accessible only via cloud services, requiring end-user devices to query a quantum cloud for real-time inference introduces unacceptable latency and cost. Our paradigm circumvents this issue by executing inference entirely on classical hardware. Consequently, the trained model can be seamlessly deployed to standard servers, laptops, or embedded devices without relying on any quantum infrastructure, thereby eliminating the primary obstacle to widespread commercialization. Furthermore, this approach strategically concentrates the quantum advantage exclusively on the training phase. While model training is fundamentally computationally intensive---making it uniquely suited to leverage the high-dimensional parallelism and entanglement properties of quantum states as powerful regularizers---inference is a relatively lightweight process. Once parameters are fixed, forward propagation can be executed rapidly and cost-effectively by classical processors. By restricting expensive quantum computational resources to the one-time training phase and delegating ubiquitous, distributed inference tasks to inexpensive classical hardware, this strategy provides a highly efficient and scalable economic model for QML applications.

Future research directions are multifaceted. First, to further improve the expressive capacity of the QCBM, more advanced circuit architectures could be investigated. For instance, deeper circuit structures, more expressive entanglement schemes, or hardware-aware configurations may enable improved modeling of complex higher-order dependencies in financial time series. Second, alternative methodologies for fusing quantum prior information with classical models can be explored, such as directly concatenating the prior embeddings from the QCBM with the input feature embeddings of the LSTM model. It is worth noting, however, that according to the backpropagation algorithm, integrating the QCBM module closer to the front-end (i.e., input layer) of the classical network may restrict the scope of its gradient influence. Consequently, early fusion strategies might constrain the effectiveness of the quantum module in guiding the global representation learning, potentially impacting the overall efficacy of the implicit knowledge distillation and quantum-assisted training processes. Third, the \textit{``quantum-assisted training with classical-efficient inference''} paradigm introduced in this work offers significant opportunities for expansion. Our current approach, which leverages a QCBM as a quantum regularizer to achieve implicit knowledge distillation, is fundamentally based on a parallel fusion architecture between the quantum and classical modules. In contrast, future efforts could explore explicit quantum distillation techniques tailored for sequential fusion architectures. For example, mature sequential hybrid quantum-classical models \cite{bharti2022noisy, cerezo2021variational,havlivcek2019supervised,lloyd2018quantum,mitarai2018quantum} could be utilized as teacher networks to explicitly guide the training of purely classical student models. This would further minimize the demand for quantum resources during deployment while striving to retain quantum-induced performance gains. Furthermore, the application potential of this hybrid framework extends far beyond volatility forecasting. It can be generalized to other key financial tasks, such as credit scoring and fraud detection, by learning the intrinsic distributions of customer data. Finally, the hybrid quantum-classical model proposed in this paper constitutes a general framework that can potentially be extended to a broader range of machine learning domains. For instance, in computer vision, the QCBM may serve as a structured probabilistic prior to enhance representation learning, feature extraction, or anomaly detection tasks. In natural language processing, it could be employed to model complex semantic or contextual distributions, thereby supporting tasks such as sentiment analysis, sequence classification, or contextual understanding.

\bibliographystyle{unsrt}
\bibliography{references}

\end{document}